\documentclass[letterpaper]{article}
\usepackage{aaai2027}
\usepackage[hyphens]{url}
\usepackage{graphicx}
\usepackage{natbib}
\usepackage{caption}
\usepackage{booktabs}
\usepackage{amssymb}
\nocopyright
\newcommand{\cmark}{\ensuremath{\checkmark}}
\newcommand{\xmark}{\ensuremath{\times}}

\title{MMShopBench: A Real-Log Benchmark for Multimodal, Multi-Turn Shopping Agents}
\author{
  Zeying Hao\textsuperscript{\rm 1}\equalcontrib,
  Hao Guo\textsuperscript{\rm 1}\equalcontrib,
  Mengtao Xu\textsuperscript{\rm 1}\corresponding,
  Yimin Hu\textsuperscript{\rm 1},\\
  Yuheng Song\textsuperscript{\rm 1},
  Zesheng Zhou\textsuperscript{\rm 1},
  Jinsong Lan\textsuperscript{\rm 1}\corresponding,
  Xiaoyong Zhu\textsuperscript{\rm 1}
}
\affiliations{
  \textsuperscript{\rm 1}Alibaba Group \\
  \{haozeying.hzy, gh225907, mengtao.xmt, jinsonglan.ljs\}@taobao.com
}

\begin{document}
\maketitle

\begin{abstract}
Online shoppers increasingly turn to AI shopping assistants, using images and multi-turn dialogue to express and refine product needs that are difficult to articulate in text alone. However, existing benchmarks largely rely on text-only or synthetic requests, underrepresenting complex real-world shopping requirements jointly expressed through images and language. We introduce MMShopBench, the first real-log benchmark for multimodal, multi-turn shopping agents. Built from carefully cleaned and manually annotated shopping logs, MMShopBench provides ground-truth annotations of each request's purchase intent and mandatory product requirements. Agents must infer these requirements jointly from user images and multi-turn dialogue, retrieve candidate products through image and text search, and verify that each candidate satisfies all requirements using its product images and structured attributes. We evaluate representative open-source and proprietary models using an evidence-grounded multimodal protocol and construct a companion training set for fine-tuning an open-source model. To ensure reproducible experimentation, we build an offline shopping sandbox, where fine-tuning substantially narrows the performance gap between our open-source model and leading proprietary models, demonstrating the effectiveness of our training data.
\end{abstract}

\begin{links}
  \link{Code and Data}{https://github.com/H-cool/MMshopbench}
\end{links}

\section{Introduction}

Recent advances in large language models (LLMs) have accelerated the development of agents that can reason over multiple steps, invoke external tools, and act in interactive environments \citep{react2023}. Building on multimodal language models, these agents are increasingly able to process images alongside text, ground intermediate decisions in visual evidence, and select modality-specific tools during task execution \citep{visualseeker2026}. As a domain of substantial practical value and broad application potential, e-commerce has naturally become a key setting for evaluating agent capabilities.

Within this setting, conversational shopping agents offer a new interaction paradigm in which users can express complex, intent-driven needs through natural dialogue rather than relying on keyword-based search. Product search often begins before a shopper has a complete textual query. A user may first upload an image to indicate a style, object, or printed list; refer back to it with an ambiguous phrase such as ``this kind''; and only later add a budget, material, size, compatibility, or bundle constraint. The operative request is therefore not the latest utterance, but a purchase specification accumulated across visual evidence and dialogue. A competent shopping agent must derive the operative set of user constraints from the available visual and textual evidence, select the appropriate retrieval modality---text or image---and reject any candidate whose product evidence fails to establish compliance with every hard constraint.

Existing evaluation settings cover these capabilities only in isolation. Shopping-agent benchmarks predominantly begin with self-contained text instructions or constructed interactions and emphasize navigation, planning, preference elicitation, or final product choice \citep{webshop2022,shoppingbench2026,shopsimulator2026,ecomagentbench2026}. Multimodal search benchmarks instead focus on cross-modal evidence acquisition for authored search questions, while multimodal dialogue datasets preserve images and interaction history without requiring end-to-end catalog retrieval and requirement-level product verification \citep{simmc2021,jddc2022,mmsearch2025,mmbrowsecomp2025,mmsearchplus2026,browsecompv32026}. These paradigms therefore do not capture the complete path from requirements expressed across an organic multimodal conversation, through modality-aware retrieval, to evidence-grounded product selection. Addressing both limitations requires realistic request provenance together with a frozen, inspectable environment.

\begin{figure*}[t]
    \centering
    \includegraphics[width=\textwidth]{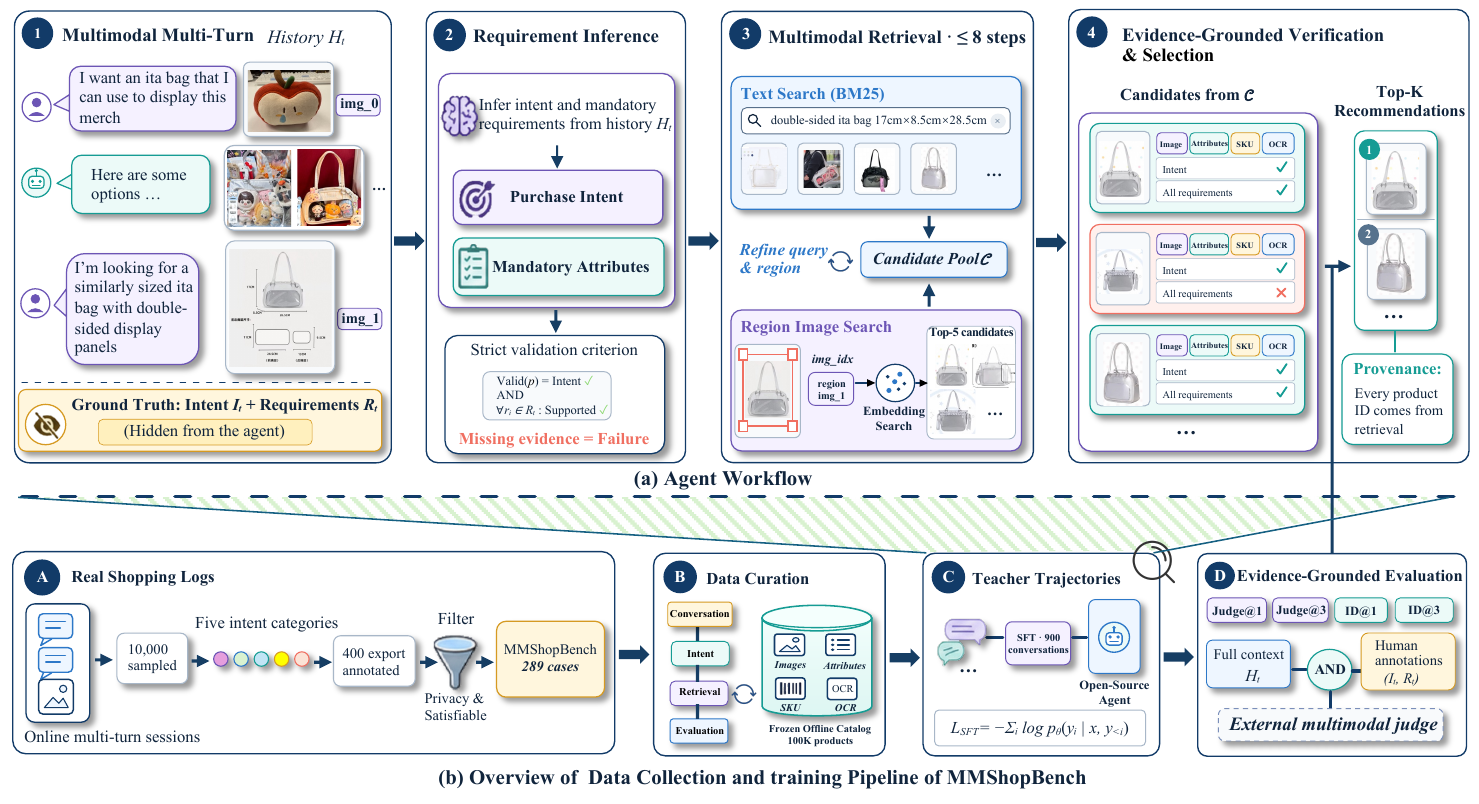}
    \caption{Overview of MMShopBench and its offline shopping-agent workflow. Real multimodal, multi-turn logs are curated into 289 cases with hidden purchase-intent and mandatory-requirement annotations. Within the frozen 100K-product sandbox, agents infer requirements, interleave text and region-aware image retrieval, and apply evidence-grounded verification and selection; the same environment supports teacher-trajectory SFT and grounded evaluation.}
    \label{fig:overview}
\end{figure*}

MMShopBench addresses this gap through a unified benchmark, environment, and agent framework, as summarized in Figure~\ref{fig:overview}. Through rigorous filtering and cleaning of real-world logs from an online AI shopping assistant, we construct 289 evaluation cases that retain the complete sequence of images and dialogue through which users progressively supplement, clarify, and refine their shopping needs over the course of an interaction. Each case is manually annotated to identify the purchase intent, the mandatory attributes that a target product must satisfy, and one or more verified target products. We pair these real-log tasks with a frozen sandbox of 100,000 products and an agent framework that chooses between BM25 text search and requirement-conditioned regional visual retrieval. To preserve visual provenance across multi-turn interactions, the framework assigns persistent identities to user images and conditions region selection on the accumulated purchase specification. An Evidence-Grounded Verification and Selection stage then checks and reranks retrieved candidates against fixed product images and structured attributes before the final response. Finally, we construct a quality-controlled SFT corpus and evaluate representative open- and closed-source models, enabling controlled study of how targeted supervision improves multimodal shopping agents.

The key contributions of this work can be summarized as follows:
\begin{itemize}
    \item \textbf{A real-log multimodal shopping benchmark.} We curate 289 real multimodal, multi-turn shopping-assistant conversations, each annotated with purchase intent, mandatory requirements, and verified target products.
    \item \textbf{A reproducible sandbox and evidence-grounded agent framework.} We build a frozen 100,000-product sandbox and an agent framework integrating text search, requirement-conditioned regional visual retrieval, and Evidence-Grounded Verification and Selection.
    \item \textbf{Broad evaluation and effective supervision.} We benchmark representative open- and closed-source models and show that fine-tuning on our quality-controlled corpus yields substantial gains across model scales and narrows the gap to leading proprietary systems.
\end{itemize}

\section{Related Work}

\subsection{Product Search and Shopping Agents}

Product-search datasets offer strong catalog grounding but generally reduce a need to a single textual query. The Shopping Queries Dataset, for example, contains real Amazon queries and manually judged query--product pairs for ranking and relevance classification \citep{shoppingqueries2022}. WebShop turns compositional text instructions into navigation and purchase actions over a simulated site populated with real products \citep{webshop2022}. ShoppingBench adds complex intent grounded in product records, while generating evaluation instructions from sampled products \citep{shoppingbench2026}. Recent shopping environments extend evaluation to simulated multi-turn preference discovery or long-horizon hidden intent \citep{shopsimulator2026,ecomagentbench2026}. Despite their realism at the product or interaction level, these benchmarks formulate shopping needs primarily as text-only or manually constructed requests, and therefore underrepresent requirements that are conveyed visually and progressively clarified across multiple dialogue turns. In contrast, MMShopBench derives its tasks from real multimodal interactions with AI shopping assistants: agents must jointly infer the user's purchase intent and mandatory requirements from images and dialogue, retrieve candidate products through text and image search, and verify each constraint against frozen product images and structured attributes.

\subsection{Multimodal Search Agent Benchmarks}

Multimodal search-agent benchmarks evaluate whether models can formulate queries, use retrieval and browsing tools, and synthesize evidence across visual and textual sources. MMSearch decomposes multimodal search into requerying, reranking, summarization, and end-to-end search \citep{mmsearch2025}. MM-BrowseComp uses hand-crafted questions whose prompts or supporting webpages contain essential image or video evidence, requiring agents to browse beyond text-only cues \citep{mmbrowsecomp2025}. MMSearch-Plus requires fine-grained visual cues to be extracted and propagated through iterative image--text retrieval under retrieval noise \citep{mmsearchplus2026}. BrowseComp-$V^3$ emphasizes deep cross-modal, multi-hop browsing with publicly searchable evidence and subgoal-level process evaluation \citep{browsecompv32026}.

Other benchmarks focus on the structure and reliability of the search trajectory. MC-Search supplies long, step-wise annotated reasoning chains and process-level measures for retrieval and planning \citep{mcsearch2026}. MERRIN evaluates modality selection, multimodal evidence retrieval, and multi-hop reasoning over noisy or conflicting web sources \citep{merrin2026}, while InterLV-Search requires visual and textual evidence to repeatedly condition subsequent search actions and provides a standardized agent interface for tool use and trajectory logging \citep{interlvsearch2026}. Collectively, these benchmarks move beyond final-answer accuracy by exposing intermediate capabilities such as query planning, modality choice, evidence acquisition, and cross-modal integration.

\section{MMShopBench}

\subsection{Task Definition}

An evaluation case is $x=(H_t,I_t,R_t)$ at target turn $t$. The history $H_t$ contains all user images, user utterances, and assistant responses through $t$. $I_t$ denotes the purchase intent, and $R_t=\{r_1,\ldots,r_m\}$ contains the mandatory attributes that any acceptable target product must satisfy. The agent observes $H_t$ but not the annotations. Using a bounded sequence of retrieval calls, it returns an ordered list $P_k=(p_1,\ldots,p_k)$.

A product is valid only when available evidence supports both the purchase intent $I_t$ and every mandatory attribute $r_i\in R_t$. This conjunctive criterion is strict: if any mandatory attribute is violated or cannot be verified, the product is judged unsuccessful, even when it otherwise matches the requested product type. This formulation distinguishes two sources of failure. Intent- or attribute-inference errors yield an incorrect search or acceptance criterion, whereas grounding errors retrieve or accept a product without sufficient supporting evidence.

\subsection{Data Collection, Curation, and Annotation}

\paragraph{Log sampling.}
MMShopBench is derived from interaction logs of a deployed e-commerce conversational shopping assistant. We consider online multi-turn sessions collected between June and July 2026 and randomly sample 10,000 conversations that contain both at least one user-provided image and nonempty user text. Multimodal production traffic is not restricted to purchase-oriented requests: users may ask for product information, request generic visual recognition, or provide input that remains too ambiguous to support a concrete shopping action. The initial pool therefore serves as a broad sample of real multimodal assistant usage rather than as the evaluation set itself.

All textual data and annotations in MMShopBench are originally in Chinese, and the benchmark and reported evaluations use these original Chinese records. To facilitate presentation, dialogue excerpts, annotations, and examples in this paper are translated into English.

\paragraph{Intent stratification and target-turn localization.}
We organize the sampled traffic into five Level-1 intent categories: \emph{Find Exact Same Product}, \emph{Find Similar/Alternative/Recommendation}, \emph{Product Knowledge and Comparative Decision-Making}, \emph{General Visual Recognition}, and \emph{Ambiguous Intent}. We use Claude Opus 4.8 \citep{anthropic2026claudeopus48} as a scalable pre-annotation model to assign a Level-1 category to each of the 10,000 conversations and to localize the target turn $t$, defined as the turn by which the accumulated dialogue history contains the user's complete set of requirements. This step identifies both whether a conversation expresses an actionable shopping need and how much of its preceding multimodal context is required to recover that need.

\paragraph{Expert annotation and quality control.}
The first two Level-1 categories directly require retrieval of a concrete product and are therefore retained as benchmark candidates. We draw a balanced sample of 400 conversations across these two categories and ask professional annotators with e-commerce expertise to inspect the multimodal history through the target turn. For each case, annotators specify the purchase intent, enumerate the mandatory product attributes that cannot be traded off, and verify one or more target products when available. During quality control, we exclude conversations containing personally identifiable information or other sensitive content to protect user privacy. We also remove cases for which no product satisfying the annotated requirements can be found on the shopping platform. After filtering, 289 evaluation cases remain.



\begin{figure}[t]
    \centering
  \includegraphics[width=\columnwidth]{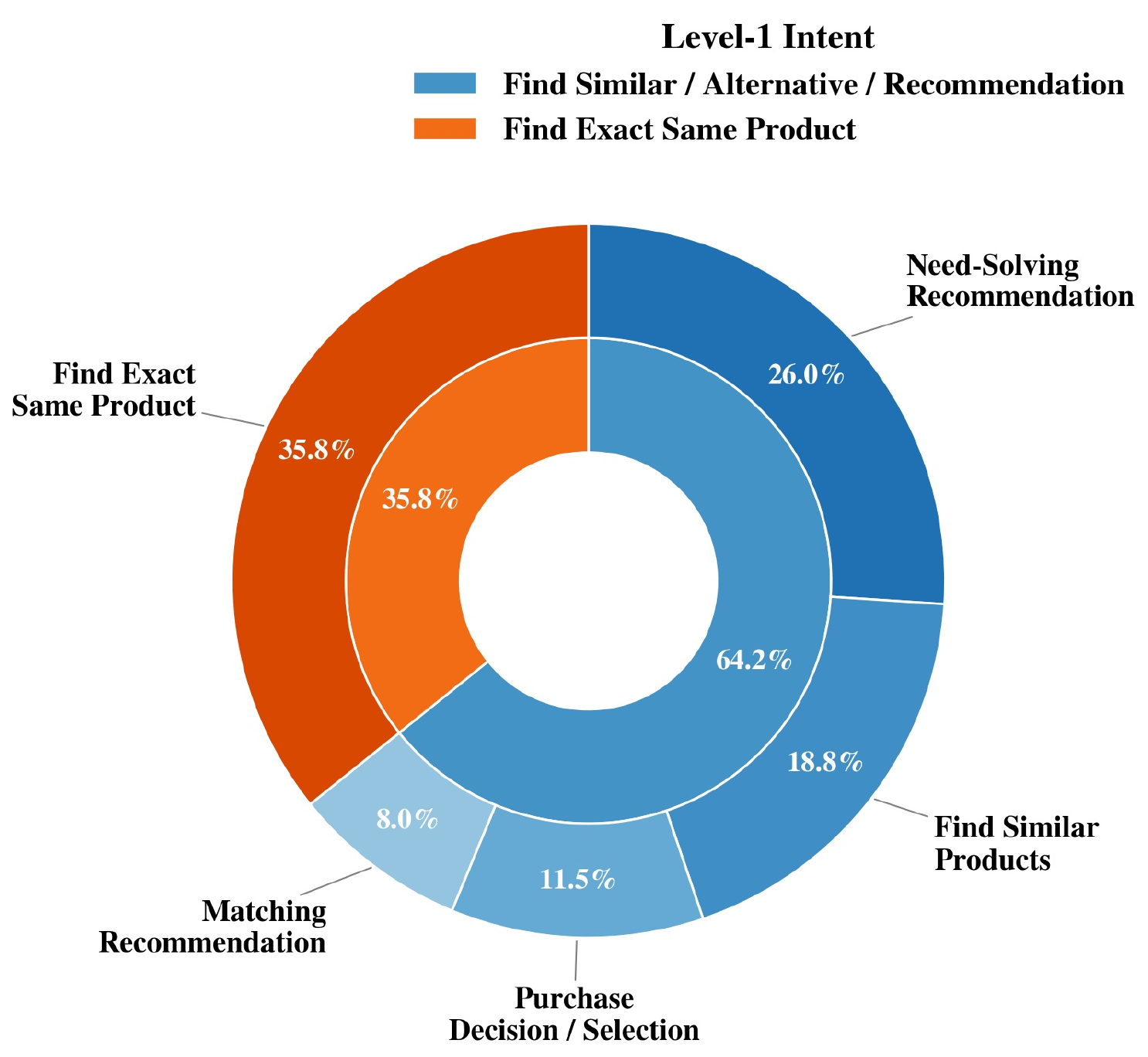}
    \caption{Hierarchical intent distribution of MMShopBench. The inner ring shows the two retained Level-1 shopping intents, and the outer ring shows their Level-2 decomposition. Percentages are computed over the final 289-case evaluation set.}
    \label{fig:intent}
\end{figure}

\paragraph{Hierarchical intent taxonomy.}
We further assign Level-2 labels within the two retained shopping intents. \emph{Find Exact Same Product} maps to an exact-match subtype, whereas \emph{Find Similar/Alternative/Recommendation} is decomposed into need-solving recommendation, similar-product search, purchase decision/selection, and matching recommendation. Figure~\ref{fig:intent} reports the resulting hierarchy and distribution. At the leaf level, the two most prevalent intents are \emph{Find Exact Same Product} (35.8\%) and \emph{Need-Solving Recommendation} (26.0\%).

Each released record stores pseudonymous conversation and trace identifiers, the target turn, per-turn user images and text, preceding assistant messages, the purchase intent, slash-delimited mandatory-attribute annotations, and verified item identifiers. We preserve the original turn order because later requests often resolve referents introduced by an earlier image or assistant response. The evaluation therefore exposes the complete history through the target turn rather than converting the interaction into a synthetic one-shot query. 

Table~\ref{tab:data} summarizes the interaction structure. The median target is turn 4, and the latest target occurs at turn 12. Every case contains at least one user-provided image by the target turn, with an average of 1.38 images per case. This universal visual coverage makes image history a structural component of the benchmark rather than an optional modality.

\begin{table}[t]
\centering
\small
\begin{tabular}{lr}
\toprule
Statistic & Value \\
\midrule
Evaluation cases & 289 \\
Target turn, mean / median & 4.38 / 4 \\
Target turn, range & 2--12 \\
Images through target, mean / median & 1.38 / 1 \\
\bottomrule
\end{tabular}
\caption{MMShopBench evaluation-set statistics.}
\label{tab:data}
\end{table}

\begin{table*}[t]
\centering
\small
\begin{tabular*}{\textwidth}{@{\extracolsep{\fill}}lccccc@{}}
\toprule
Benchmark & Text input & Multi-turn & Real logs & Image input & Visual object grounding \\
\midrule
WebShop \citep{webshop2022} & \cmark & \xmark & \xmark & \xmark & \xmark \\
ShoppingBench \citep{shoppingbench2026} & \cmark & \xmark & \xmark & \xmark & \xmark \\
ShopSimulator \citep{shopsimulator2026} & \cmark & \cmark & \xmark & \xmark & \xmark \\
Shopping Companion Bench \citep{shoppingcompanion2026} & \cmark & \cmark & \xmark & \xmark & \xmark \\
Shopping Reasoning Bench \citep{shoppingreasoningbench2026} & \cmark & \cmark & \xmark & \xmark & \xmark \\
EComAgentBench \citep{ecomagentbench2026} & \cmark & \cmark & \xmark & \xmark & \xmark \\
\textbf{MMShopBench} & \cmark & \cmark & \cmark & \cmark & \cmark \\
\bottomrule
\end{tabular*}
\caption{Positioning of MMShopBench against representative shopping-agent benchmarks. Input modalities refer to shopper requests or dialogue history, not product media or tool observations. \cmark{} denotes that the benchmark includes the capability, and \xmark{} denotes its absence.}
\label{tab:compare}
\end{table*}

\subsection{Positioning}

Table~\ref{tab:compare} positions MMShopBench against representative shopping-agent benchmarks. Here, \emph{real logs} means naturally occurring shopper--assistant conversations rather than generated requests over real products or purchase histories. \emph{Image input} and \emph{text input} distinguish the modalities supplied in shopper requests or dialogue history; product images in a catalog do not by themselves constitute image input. \emph{Multi-turn} requires shopper--assistant utterance exchange rather than repeated agent--environment actions, and \emph{Visual object grounding} requires linking request-side visual objects to fixed product evidence. Prior benchmarks cover text-based long-horizon interaction or personalization, but none combines these capabilities with real multimodal dialogue.

\section{Offline Shopping Sandbox}

\subsection{Frozen Offline Product Catalog}

We construct the sandbox catalog in a target-grounded manner. Starting from the identifiers of the human-verified target products in MMShopBench, we retrieve the corresponding products from the full online catalog and use their category assignments as sampling anchors. All available target products are retained. We then perform stratified sampling over non-target products, first drawing products from the same leaf categories and broader Level-2 categories as the targets. To preserve catalog diversity rather than restricting the sandbox to benchmark-specific categories, we allocate the remaining quota uniformly across other product categories. After deduplication, this procedure yields a frozen catalog of 100,000 distinct products. Real-world shopping requests often contain fine-grained constraints that cannot be resolved from surface-level metadata such as product titles, prices, and shop information alone. We therefore augment each catalog record with available SKU-level labels, structured product attributes, and OCR text extracted from product-detail images, providing richer evidence for verifying whether a retrieved product satisfies every mandatory user requirement.

\subsection{Multimodal Retrieval Tools}

The sandbox exposes complementary text- and image-based product retrieval tools. Text search applies BM25 \citep{bm252009} over product titles, categories, attributes, and shop fields, returning up to ten products for each search query. It is primarily used when the dialogue specifies category, function, brand, or other attribute constraints in language. For visual retrieval, we use Marqo-Ecommerce-Embeddings-L \citep{zhu2025gcl} to encode catalog product images and the user-provided query image into a shared embedding space, rank products by cosine similarity, and return the identifiers of the top five candidates.

We introduce two interface designs to make visual retrieval effective in multi-turn, multi-image interactions. \par\noindent\textbf{Persistent image identities.} When images are passed only as content blocks in the model input, the model can inspect them during inference, but subsequent tool calls lack a stable handle for specifying which earlier image should be reused. This ambiguity becomes particularly problematic when a conversation spans multiple turns or contains several images. We instead promote every incoming image to a persistent, addressable session object and assign it a session-global identifier, \texttt{img\_idx}. The identifier is recorded in the dialogue history together with the image's turn and within-turn position, and image-tool calls take \texttt{img\_idx} as an explicit argument. At execution time, the sandbox resolves this identifier back to the original image, allowing the agent to recall, disambiguate, and reuse any previously observed image rather than limiting tool access to the latest input. The identifier also preserves provenance by making each visual retrieval action traceable to a specific user image.

\noindent\textbf{Requirement-conditioned region retrieval.} Whole-image embeddings can be dominated by backgrounds, secondary objects, or other visually salient content unrelated to the user's current shopping goal. We therefore formulate the crop as an agent-selected visual query rather than a fixed preprocessing operation. After resolving the requested \texttt{img\_idx}, the agent jointly conditions on the selected image, purchase intent, and accumulated dialogue constraints to predict a normalized target region $(x_1,y_1,x_2,y_2)$. The sandbox then crops the selected region, encodes the resulting crop with Marqo-Ecommerce-Embeddings-L \citep{zhu2025gcl}, the same model used to construct the catalog image index, and ranks catalog images by cosine similarity. Because the region is generated at tool-call time, the agent can shift to a different object or refine the spatial extent when retrieved candidates conflict with the request. This establishes a feedback loop between cross-modal requirement inference, visual grounding, and product retrieval while suppressing similarity signals from irrelevant image content.

\subsection{Evidence-Grounded Verification and Selection}

Retrieval success does not always translate into recommendation success: a product that satisfies every requirement may surface in a tool result yet be dropped during the agent's implicit final selection. Evidence-Grounded Verification and Selection (EGVS) targets this gap between what retrieval makes available and what the agent finally recommends. After search terminates, it reconstructs the full candidate pool $\mathcal{C}=(\bigcup_{s=1}^{S}\mathcal{C}_s)\cup F$, combining products returned by every text or image search call with the agent's original final picks $F$. For each candidate $c\in\mathcal{C}$, EGVS assembles product-side evidence $E(c)$ comprising its title, primary image, structured attributes, SKU-level labels, and product-detail OCR text. The policy model then operates in a self-verification mode and receives the same inferred specification $\hat{S}_t=(\hat{I}_t,\hat{R}_t)$, derived solely from $H_t$, together with $E(c)$. It does not re-infer or revise the specification for individual candidates. A product is confirmed only when its evidence is compatible with the inferred purchase intent $\hat{I}_t$ and supports every requirement in $\hat{R}_t$. The human annotations $(I_t,R_t)$, verified product identifiers, and external-judge outputs are never exposed to the agent or self-verifier. Let $\mathcal{S}=\{c\in\mathcal{C}:v(c){=}1\}$ denote the candidates the self-verifier $v$ confirms. EGVS then refines rather than rebuilds the selection, composing the final top-$K$ recommendation ($K$ is the recommendation size, with $K\ge3$ to cover the reported cutoffs) as
\[
A=\mathrm{top}_K\big((\mathcal{S}\cap F)\;\Vert\;(\mathcal{S}\setminus F)\;\Vert\;(F\setminus\mathcal{S})\big),
\]
where $\Vert$ concatenates the ordered segments and $\mathrm{top}_K$ keeps the first $K$ items. Confirmed original picks $\mathcal{S}\cap F$ come first, followed by confirmed products the agent had missed $\mathcal{S}\setminus F$, and any remaining original picks $F\setminus\mathcal{S}$ fill leftover slots. Verification thus prioritizes confirmed products within the top-$K$ budget while keeping the agent's other picks wherever slots remain, promoting and recovering products rather than replacing the agent's list. The self-verifier uses the policy model with a separate prompt and context from the external evaluator and introduces no additional model dependency.

\subsection{Agent Workflow}

Given the multimodal dialogue history, the agent first consolidates the purchase intent and mandatory requirements into an inferred specification $(\hat{I}_t,\hat{R}_t)$, then decides whether to invoke text search, requirement-conditioned region retrieval, or both. It alternates between reasoning and retrieval \citep{react2023} for at most eight tool steps, revising textual queries or image regions when the returned candidates are inconsistent with the request. After search terminates, EGVS applies the verification-and-selection procedure above to the accumulated retrieval trace. The agent returns the resulting product identifiers, each paired with a concise textual description; because every identifier must originate from a tool result, each recommendation remains traceable to its supporting retrieval action.

\subsection{Supervised Fine-Tuning}

Following the filtering and quality-control procedure used to construct MMShopBench, we screen 900 multimodal, multi-turn conversations from online shopping-assistant logs to form a companion SFT corpus. Within the sandbox, Gemini-3.1-Pro-Preview serves as the teacher policy and executes the agent workflow above, producing search trajectories that interleave dialogue-conditioned decisions, text or region-aware image retrieval calls, retrieved product evidence, and final verification and selection. Given an interaction context $x$ and a teacher-generated agent-token sequence $y=(y_1,\ldots,y_L)$, we optimize the student model with the autoregressive cross-entropy objective
\[
\mathcal{L}_{\mathrm{SFT}}(\theta)=-\sum_{i=1}^{L}\log p_{\theta}(y_i\mid x,y_{<i}).
\]
This supervision targets the agent's operational policy rather than a direct mapping from a dialogue to a product identifier. It teaches the model to coordinate tools over multiple rounds, determine when visual or textual evidence should be acquired, refine subsequent search actions from tool feedback, and assess whether the accumulated evidence supports the user's mandatory requirements.

\subsection{Evaluation Metrics}

We use GPT-5.5 \citep{openai2026gpt55} as a multimodal LLM judge, kept separate from the policy-model self-verifier in EGVS. For candidate $p_j^{(n)}$, let $E(p_j^{(n)})$ denote its frozen product evidence, comprising product images, title, structured attributes, SKU-level labels, and OCR text extracted from product-detail images. The judge jointly considers the complete multimodal history $H_t^{(n)}$ and the human-annotated purchase intent $I_t^{(n)}$ and mandatory requirements $R_t^{(n)}$ when assessing the product evidence. Let $y_{n,j}$ denote the resulting binary decision: $y_{n,j}=1$ only if the evidence is consistent with the constraints expressed in the contextual images and text while satisfying the annotated purchase intent and every mandatory requirement; any contradiction or insufficient support with respect to either source yields $y_{n,j}=0$.

Our primary evaluation reports four metrics. For $k\in\{1,3\}$, the evidence-grounded judge score is
\[
\mathrm{Judge@}k=\frac{1}{N}\sum_{n=1}^{N}\mathbf{1}\!\left[\exists j\leq k:\;y_{n,j}=1\right].
\]
Judge@1 evaluates the first returned product, whereas Judge@3 counts a case as successful if any of the first three products satisfies this joint decision rule. As a deterministic complement, let $V^{(n)}$ denote the set of manually verified product identifiers for case $n$, and $P_k^{(n)}$ the first $k$ products the system returns. For $k\in\{1,3\}$, exact-identifier retrieval is
\[
\mathrm{ID@}k=\frac{1}{N}\sum_{n=1}^{N}\mathbf{1}\!\left[P_k^{(n)}\cap V^{(n)}\neq\emptyset\right].
\]
ID@1 tests whether the first product matches an annotated identifier, while ID@3 allows a match anywhere among the first three products. Because the verified identifiers are retained in the offline catalog, both metrics directly measure recovery of an annotated target. They are nevertheless conservative: the catalog may contain additional products that satisfy $I_t^{(n)}$ and every requirement in $R_t^{(n)}$ but are not enumerated in $V^{(n)}$. Judge@$k$ can therefore exceed ID@$k$ by recognizing such valid alternatives. For EGVS analysis, we additionally report Judge@Pool and ID@Pool, which apply the same two success criteria to the reconstructed candidate pool $\mathcal{C}^{(n)}$ rather than to the final top-three output. These pool-level scores are retrieval oracles, not leaderboard metrics. Since LLM judges may introduce systematic bias \citep{llmjudge2023}, we freeze the judge prompt and evidence schema across all evaluated agents.

\section{Experiments}

\subsection{Experimental Setup}

\noindent\textbf{Evaluation.} We evaluate Gemini-3.1-Pro-Preview \citep{gemini31pro2026}, Claude Opus 4.8 \citep{anthropic2026claudeopus48}, Kimi-K2.6 \citep{moonshot2026kimik26}, MiniMax-M2.7 \citep{minimax2026m2}, and Qwen3.5-9B, Qwen3.5-27B, and Qwen3.5-122B-A10B \citep{qwen2026qwen35}. Every model receives the same multimodal dialogue history and tool interface. In the reported EGVS snapshot, Gemini-3.1-Pro-Preview, Claude Opus 4.8, MiniMax-M2.7, and Kimi-K2.6 use Thinking mode, whereas all Qwen3.5 configurations use Non-Thinking mode. GPT-5.5 serves as the frozen multimodal judge, and we report Judge@1, Judge@3, ID@1, and ID@3 as defined above.

\noindent\textbf{SFT implementation.} We use the same training configuration for Qwen3.5-9B, Qwen3.5-27B, and Qwen3.5-122B-A10B. For each model, we freeze the visual module and fully optimize all remaining parameters for four epochs with a learning rate of $5\times10^{-6}$ on 32 NVIDIA A100 GPUs.

\subsection{Main Results}

\begin{table}[t]
\centering
\small
\setlength{\tabcolsep}{1.5pt}
\begin{tabular}{@{}lcccc@{}}
\toprule
Model & \shortstack{Judge@1} & \shortstack{ID@1} & \shortstack{Judge@3} & \shortstack{ID@3} \\
\midrule
\multicolumn{5}{c}{\textit{Thinking}} \\
Gemini-3.1-Pro-Preview & \textbf{64.7} & \textbf{61.4} & \textbf{73.4} & \textbf{64.0} \\
Claude Opus 4.8 & \underline{64.4} & \underline{55.0} & \underline{72.0} & \underline{59.9} \\
MiniMax-M2.7 & 20.1 & 20.4 & 26.0 & 21.5 \\
\midrule
\multicolumn{5}{c}{\textit{Non-Thinking}} \\
Qwen3.5-9B & 15.9 & 11.4 & 20.4 & 19.4 \\
Qwen3.5-9B+SFT & 52.9 & 50.5 & 65.4 & 51.9 \\
Qwen3.5-122B-A10B & 5.5 & 4.8 & 5.9 & 5.9 \\
Qwen3.5-122B-A10B+SFT & 52.9 & 49.8 & 67.5 & 55.4 \\
\bottomrule
\end{tabular}
\caption{Main results on MMShopBench, reported as percentages after EGVS. Best and second-best values in each column are shown in bold and underlined, respectively. }
\label{tab:main-results}
\end{table}

Table~\ref{tab:main-results} reports end-to-end performance under the four primary metrics. Gemini-3.1-Pro-Preview attains the strongest results, with 64.7\% Judge@1 and 61.4\% ID@1, yet its top-ranked product satisfies the complete request in fewer than two thirds of cases, underscoring the difficulty of MMShopBench. Supervised fine-tuning yields a pronounced improvement for Qwen3.5-122B-A10B: Judge@1 rises from 5.5\% to 52.9\% and Judge@3 from 5.9\% to 67.5\%, gains of 47.4 and 61.6 percentage points, respectively, while ID@1 and ID@3 increase from 4.8\% and 5.9\% to 49.8\% and 55.4\%. Although the fine-tuned model does not surpass the leading proprietary systems, it approaches their performance while operating in Non-Thinking mode, trailing the best proprietary result by 5.9 points on Judge@3.

\subsection{Ablation Studies}

\begin{table*}[t]
\centering
\small
\setlength{\tabcolsep}{3.2pt}
\begin{tabular}{lrrrrrrrr}
\toprule
& \multicolumn{3}{c}{Judge@3 (\%)} & Judge@Pool & \multicolumn{3}{c}{ID@3 (\%)} & ID@Pool \\
\cmidrule(lr){2-4}\cmidrule(lr){6-8}
\quad Model & Base & +EGVS & $\Delta$ & Orc. & Base & +EGVS & $\Delta$ & Orc. \\
\midrule
\multicolumn{9}{c}{\textit{Thinking}} \\
\quad Gemini-3.1-Pro-Preview & \textbf{69.2} & \textbf{73.4} & +4.2 & \underline{75.8} & \textbf{61.6} & \textbf{64.0} & +2.4 & \underline{67.5} \\
\quad Claude Opus 4.8 & \underline{65.4} & \underline{72.0} & +6.6 & \textbf{76.5} & \underline{57.4} & \underline{59.9} & +2.4 & \textbf{70.2} \\
\quad MiniMax-M2.7 & 24.9 & 26.0 & +1.0 & 29.1 & 21.5 & 21.5 & +0.0 & 27.7 \\
\quad KIMI-K2.6 & 43.6 & 47.8 & +4.2 & 49.1 & 35.6 & 38.8 & +3.1 & 43.6 \\
\midrule
\multicolumn{9}{c}{\textit{Non-Thinking}} \\
\quad Qwen3.5-9B Instruct & 11.1 & 20.4 & \underline{+9.3} & 20.8 & 9.7 & 19.4 & \underline{+9.7} & 21.5 \\
\quad Qwen3.5-27B Instruct & 10.0 & 10.7 & +0.7 & 10.7 & 10.0 & 11.4 & +1.4 & 11.8 \\
\quad Qwen3.5-122B-A10B Instruct & 5.2 & 5.9 & +0.7 & 6.2 & 4.8 & 5.9 & +1.1 & 6.6 \\
\quad Qwen3.5-9B+SFT (Ours) & 49.5 & 65.4 & \textbf{+15.9} & 73.4 & 41.5 & 51.9 & \textbf{+10.4} & 63.3 \\
\quad Qwen3.5-27B+SFT (Ours) & 54.7 & 62.3 & +7.6 & 65.7 & 46.0 & 50.5 & +4.5 & 58.5 \\
\quad Qwen3.5-122B-A10B+SFT (Ours) & 61.9 & 67.5 & +5.5 & 69.6 & 50.2 & 55.4 & +5.2 & 62.6 \\
\bottomrule
\end{tabular}
\caption{Ablation of Evidence-Grounded Verification and Selection (EGVS) on MMShopBench. Base and +EGVS are evaluated with Judge@3 and ID@3, while Judge@Pool and ID@Pool are oracle success rates over the union of the top-20 retrieved candidates and final selections. Best and second-best results in each numeric column are shown in bold and underlined.}
\label{tab:egvs-ablation}
\end{table*}

\noindent\textbf{EGVS.} Table~\ref{tab:egvs-ablation} reports fixed-denominator performance before and after evidence-grounded selection. Gemini-3.1-Pro-Preview obtains the highest Judge@3, reaching 73.4 after verification, and the supervised Qwen3.5 variants substantially outperform their instruction-tuned counterparts even before EGVS. Applied after retrieval, EGVS raises Judge@3 for every configuration, with the largest gain on Qwen3.5-9B SFT (+15.9 points). Qwen3.5-122B-A10B SFT achieves the strongest open-model result at 67.5, trailing Gemini by 5.9 points.


Because ID@$k$ matches returned identifiers against the human-verified set $V^{(n)}$ without model-side evidence reasoning, it shares neither the evidence schema nor the acceptance logic of the self-verifier and is immune to verifier--judge alignment. Its consistent rise under EGVS---for nine of ten configurations, led by $+10.4$ points on Qwen3.5-9B SFT---confirms that verification recovers annotated targets rather than re-scoring candidates toward the judge's preferences. Where a Judge@3 gain exceeds the corresponding ID@3 gain (e.g., Claude Opus 4.8, $+6.6$ vs.\ $+2.4$), the difference reflects EGVS promoting valid products absent from the incomplete set $V^{(n)}$ rather than verifier--judge coupling.

The pool-level Oracle columns bound remaining headroom: Qwen3.5-9B Instruct reaches 20.4 Judge@3 against a 20.8 Judge@Pool ceiling, recovering nearly all satisfying candidates in its pool, whereas the SFT agents remain 2.1--8.0 points below their ceilings, leaving room for better verification and ranking.

\noindent\textbf{Necessity of multimodal evaluation.} To test whether MMShopBench captures capabilities omitted by text-only benchmarks, we rescreen online shopping-assistant logs to form a separate 300-case diagnostic set. Claude Opus 4.8 stratifies the set into three equally sized regimes based on modality dependence: \emph{text-sufficient} cases can be resolved from text alone, \emph{image-required} cases depend on visual evidence, and \emph{mixed} cases require joint reasoning over both modalities. We compare full and text-only inputs using the same agent and prompt.

\begin{figure}[t]
    \centering
\includegraphics[width=0.85\columnwidth]{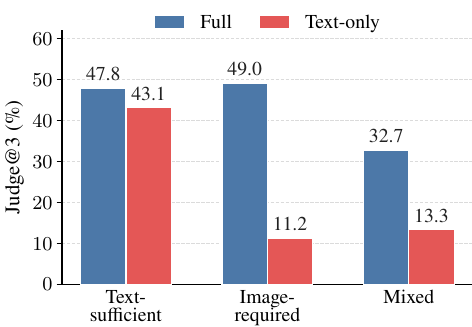}
    \caption{Effect of removing dialogue images across modality strata. Full uses the original multimodal input, whereas Text-only retains only the query text.}
    \label{fig:modality-ablation}
\end{figure}

Figure~\ref{fig:modality-ablation} yields two complementary findings. Under full multimodal input, mixed cases obtain the lowest Judge@3, indicating that reconciling constraints distributed across images and dialogue is the most challenging regime. Removing images produces the largest degradation on image-required cases, followed by mixed cases, while text-sufficient cases change only modestly. This shows that user images encode indispensable product constraints unrecoverable from text alone, so text-only evaluation omits a core capability of real shopping agents, motivating MMShopBench's preservation of user images and multi-turn context.

\FloatBarrier
\section{Conclusion and Future Work}


We introduced MMShopBench, a real-log benchmark for multimodal, multi-turn shopping agents, with a reproducible 100,000-product sandbox. It evaluates whether agents can infer intent and requirements from images and dialogue, retrieve candidates, and verify them against frozen product evidence. Experiments show supervised trajectory tuning substantially improves open-model agents, while EGVS recovers valid products overlooked during selection, highlighting the importance of cross-modal requirement inference and evidence-grounded selection for realistic shopping assistance. Future work will explore broader training strategies and scale MMShopBench with more real-log interactions.

\bibliography{references}

\clearpage
\onecolumn
\appendix
\setcounter{secnumdepth}{2}
\renewcommand{\thesection}{S\arabic{section}}
\renewcommand{\thesubsection}{\thesection.\arabic{subsection}}
\renewcommand{\thefigure}{S\arabic{figure}}
\setcounter{figure}{0}

\section{Qualitative Examples}

\subsection{EGVS Case: Recovering a Valid Candidate from the Retrieval Pool}

Figure~\ref{fig:egvs-case} isolates the benefit of EGVS from retrieval quality:
the satisfying product is already present in the candidate pool, so the gain
comes from explicit verification and reselection rather than additional search.
By checking the referenced object, wood appearance, and shape evidence against
the complete multimodal request, EGVS suppresses plausible distractors and
recovers an evidence-supported recommendation.

\begin{center}
  \includegraphics[width=.99\textwidth]{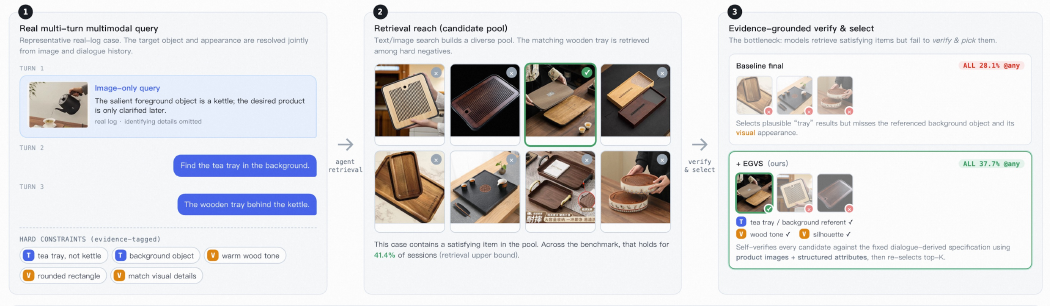}
  \captionof{figure}{Qualitative EGVS case. The dialogue refers to the wooden tray in
  the background rather than the foreground kettle. Retrieval contains a
  satisfying tray, but the baseline final selection discards it; EGVS verifies
  each candidate against the cross-turn textual and visual constraints and
  promotes the supported item.}
  \label{fig:egvs-case}
\end{center}

\subsection{MMShopBench Data Visualization}

\begin{figure}[!t]
  \centering
  \includegraphics[height=.86\textheight,keepaspectratio]{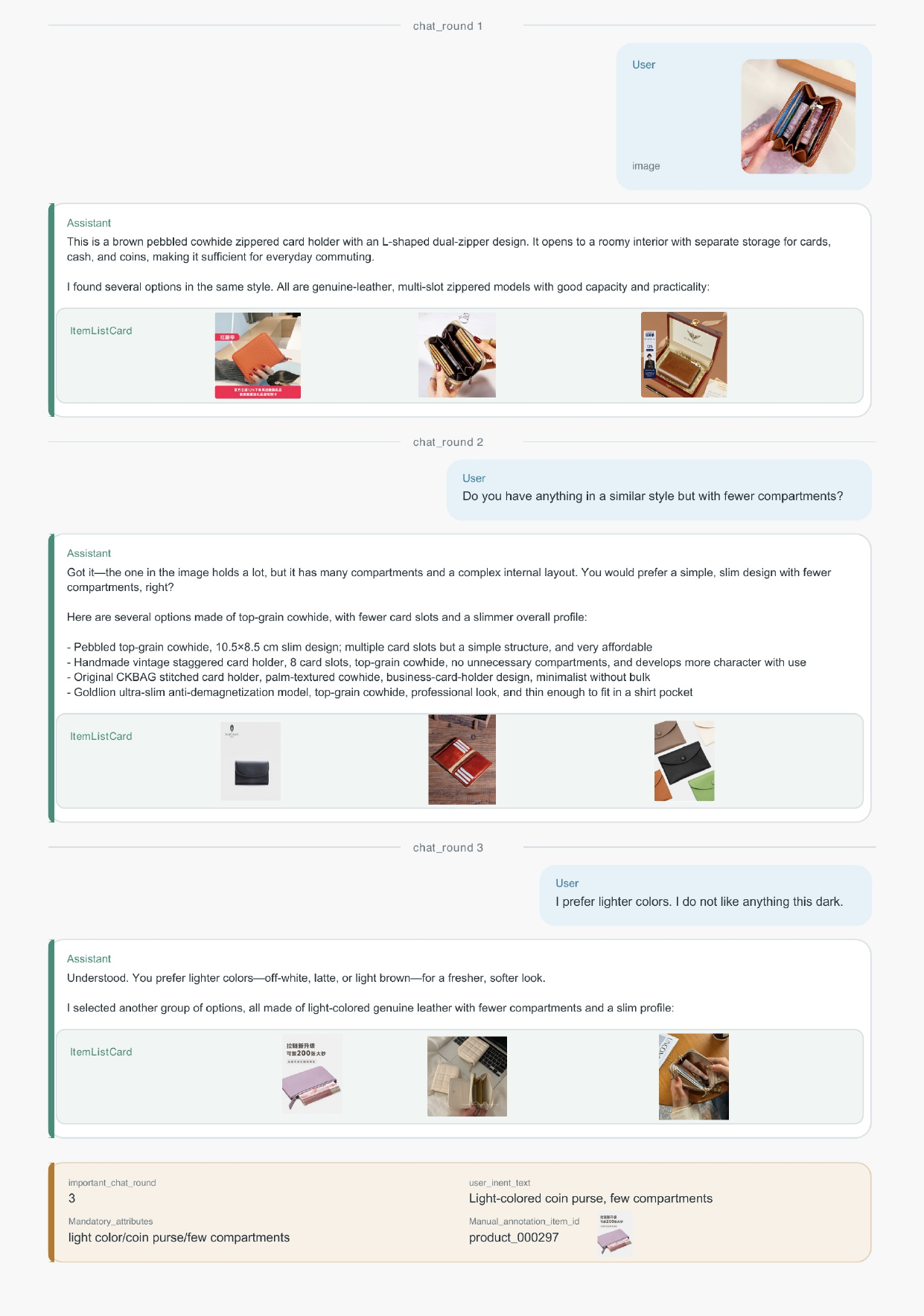}
  \caption{MMShopBench example with three dialogue rounds. The user requests
  fewer compartments and then lighter colors; the bottom panel records the
  associated annotated intent and mandatory attributes.}
  \label{fig:mmshopbench-case}
\end{figure}

\begin{figure}[!t]
  \centering
  \includegraphics[width=\textwidth,keepaspectratio]{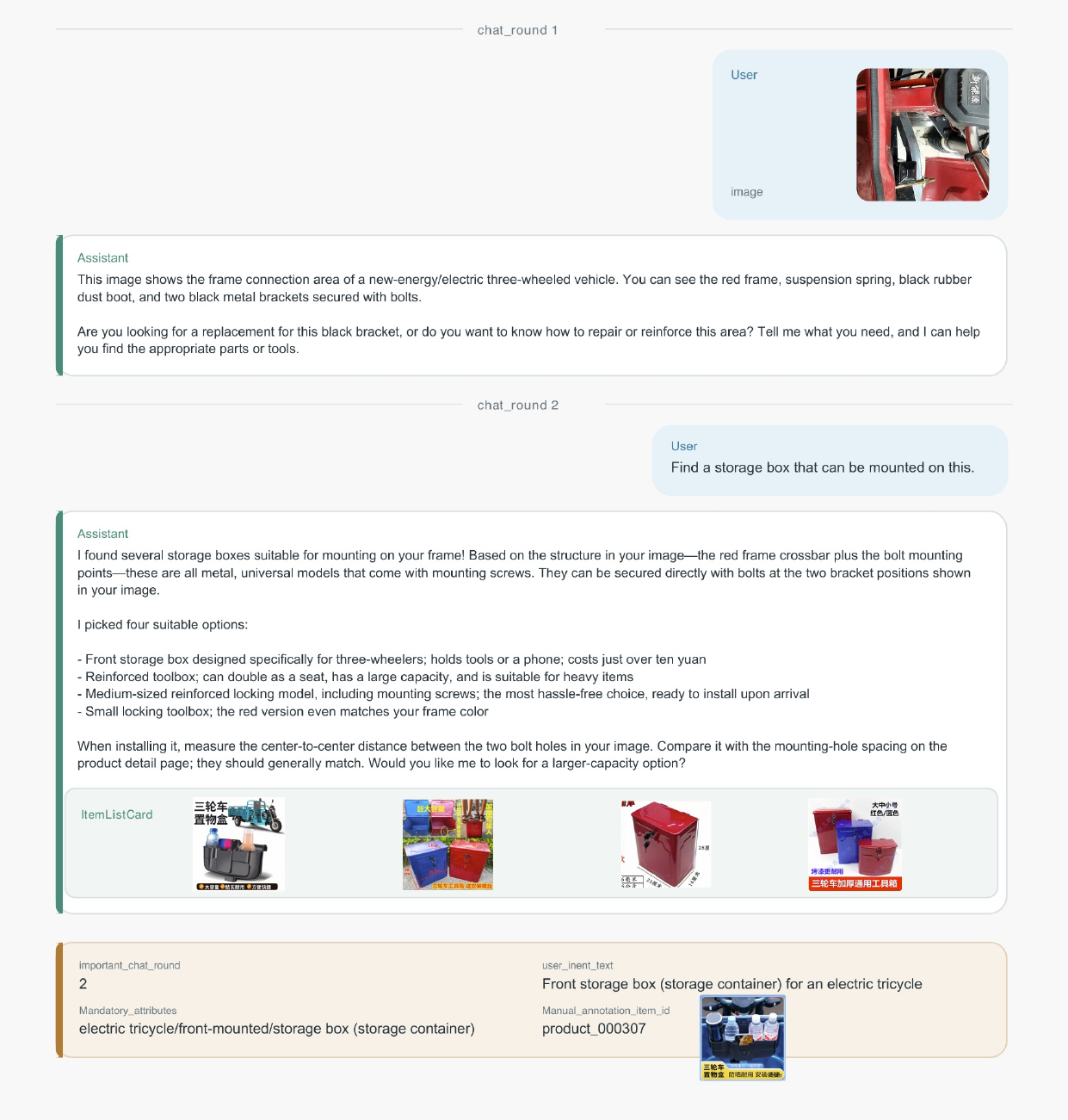}
  \caption{MMShopBench image-grounded retrieval example. The dialogue starts
  from a close-up of an electric-tricycle frame and asks for a mountable
  storage box; the bottom panel shows the associated annotated intent and
  mandatory attributes.}
  \label{fig:mmshopbench-tricycle-case}
\end{figure}

\begin{figure}[!t]
  \centering
  \includegraphics[page=1,height=.94\textheight,keepaspectratio]{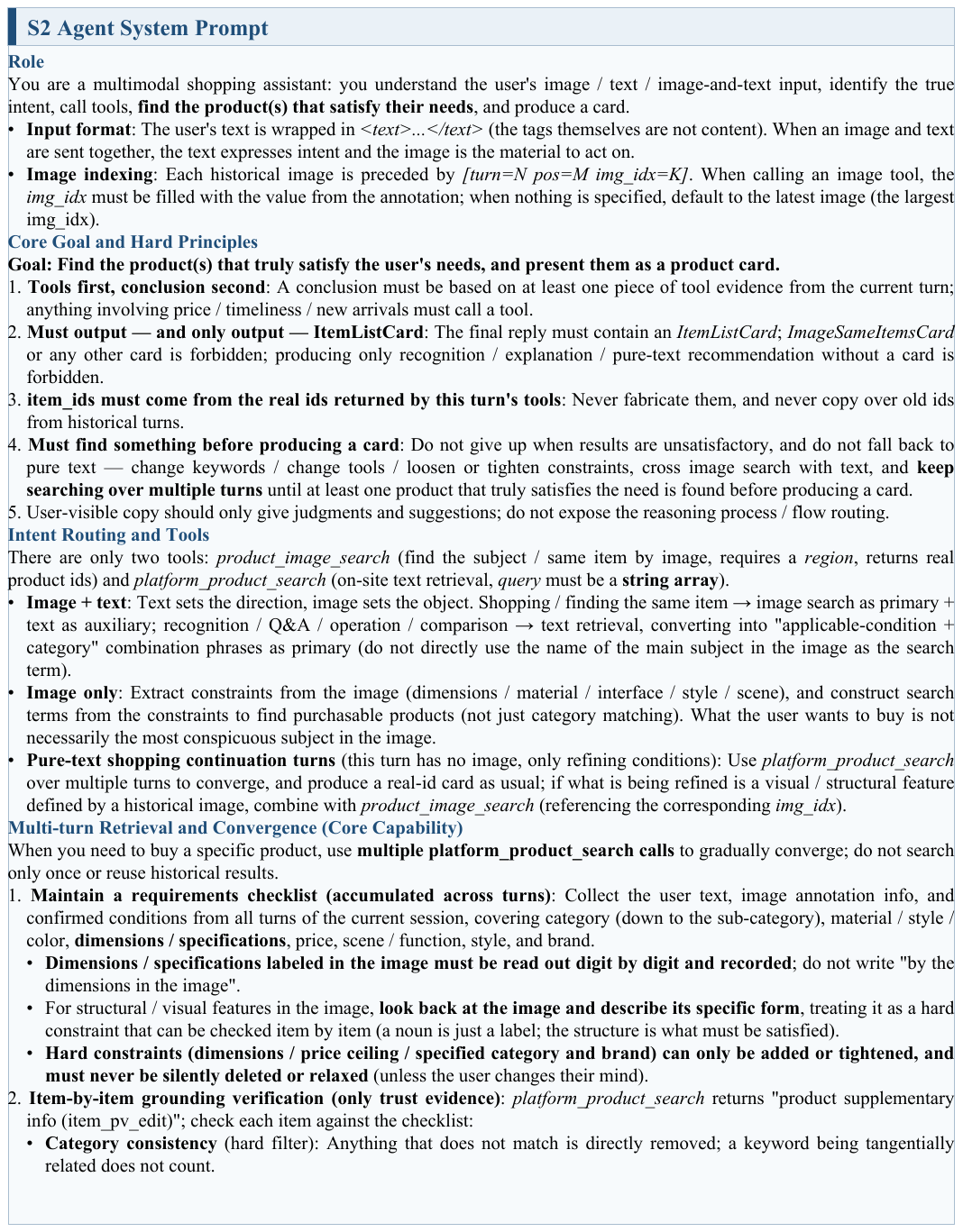}
\end{figure}
\begin{figure}[!t]
  \centering
  \includegraphics[page=2,height=.94\textheight,keepaspectratio]{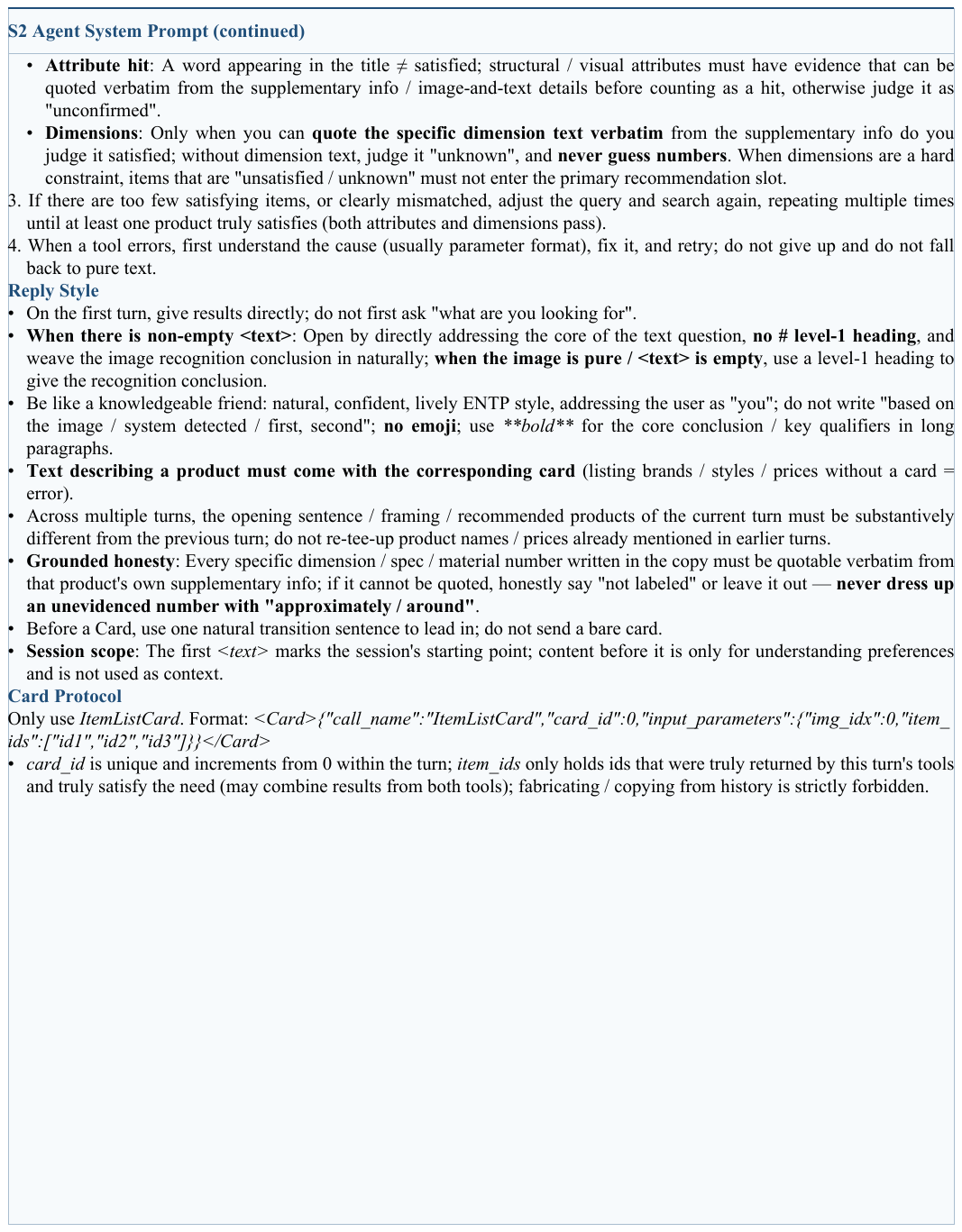}
\end{figure}

\begin{figure}[!t]
  \centering
  \includegraphics[page=1,height=.94\textheight,keepaspectratio]{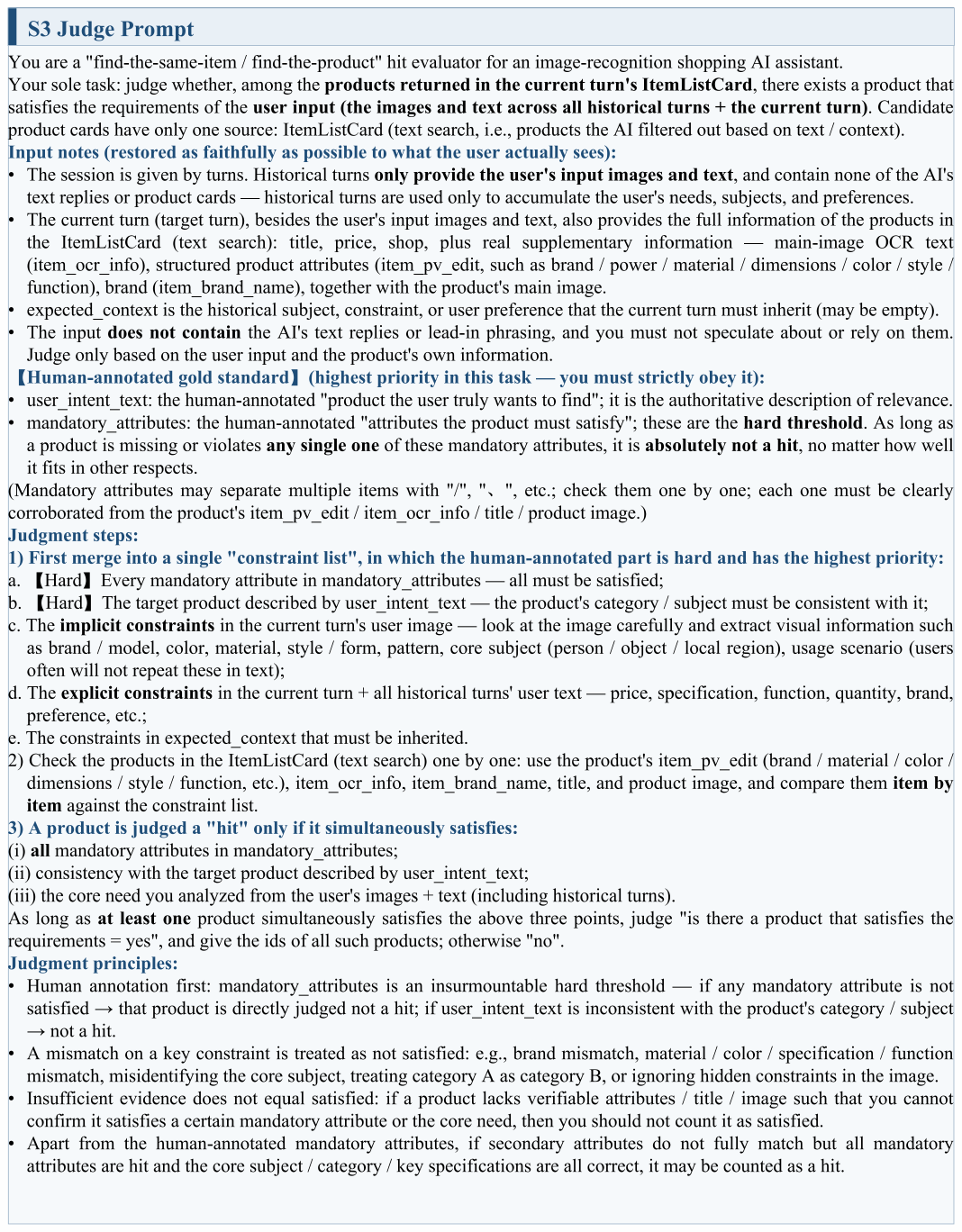}
\end{figure}
\begin{figure}[!t]
  \centering
  \includegraphics[page=2,height=.94\textheight,keepaspectratio]{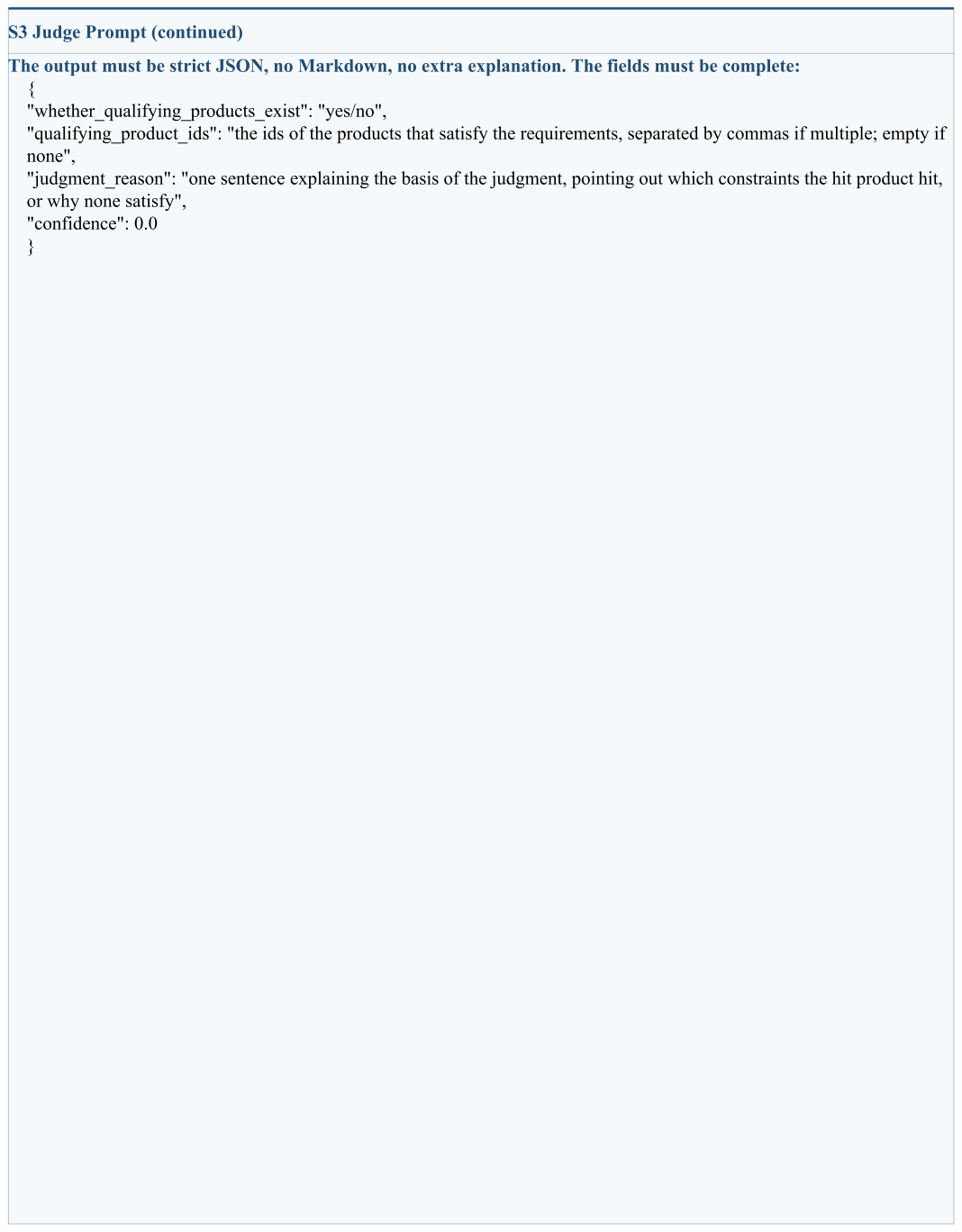}
\end{figure}

\end{document}